\documentclass[conference]{IEEEtran}
\IEEEoverridecommandlockouts
\usepackage{amsmath,amssymb,amsfonts}
\usepackage{graphicx}
\usepackage{textcomp}
\usepackage{xcolor}
\def\BibTeX{{\rm B\kern-.05em{\sc i\kern-.025em b}\kern-.08em
    T\kern-.1667em\lower.7ex\hbox{E}\kern-.125emX}}
\usepackage{latexsym}
\usepackage{stmaryrd}
\usepackage{algorithm}
\usepackage[noend]{algpseudocode}

\usepackage[caption=false,font=footnotesize]{subfig}
\usepackage[noadjust]{cite}
\usepackage{booktabs}

\begin{document}

\title{An Empirical Evaluation of the t-SNE Algorithm for Data Visualization in Structural Engineering}


\author{\IEEEauthorblockN{Parisa Hajibabaee}
\IEEEauthorblockA{\textit{Computer Science} \\
\textit{University of Massachusetts}\\
Lowell, USA \\
parisa\_hajibabaee@student.uml.edu}
\and
\IEEEauthorblockN{Farhad Pourkamali-Anaraki}
\IEEEauthorblockA{\textit{Computer Science} \\
\textit{University of Massachusetts}\\
Lowell, USA \\
farhad\_pourkamali@uml.edu}
\and
\IEEEauthorblockN{Mohammad Amin Hariri-Ardebili}
\IEEEauthorblockA{\textit{Civil Environmental and Architectural Engineering} \\
\textit{University of Colorado}\\
Boulder, USA \\
mohammad.haririardebili@colorado.edu}
\thanks{This paper has been accepted for publication in IEEE International Conference on Machine Learning and Applications 2021. Personal use of this material is permitted. Permission from IEEE must be obtained for all other uses.}

}


\maketitle

\begin{abstract}
A fundamental task in machine learning involves visualizing high-dimensional data sets that arise in high-impact application domains. When considering the context of large imbalanced data, this problem becomes much more challenging. In this paper, the
t-Distributed Stochastic Neighbor Embedding (t-SNE) algorithm is used to reduce the dimensions of an earthquake engineering related data set for visualization purposes. Since imbalanced data sets greatly affect the accuracy of classifiers, we employ Synthetic Minority Oversampling Technique (SMOTE) to tackle the imbalanced nature of such data set. We present the result obtained from t-SNE and SMOTE and compare it to the basic approaches with various aspects. Considering four options and six classification algorithms, we show that using t-SNE on the imbalanced data and SMOTE on the training data set, neural network classifiers have promising results without sacrificing accuracy. Hence, we can transform the studied scientific data into a two-dimensional (2D) space, enabling the visualization of the classifier and the resulting decision surface using a 2D plot.
\end{abstract}


\section{Introduction}\label{sec:into}

Visualizing high-dimensional data is among the most fundamental tasks in modern machine learning to broaden its reach in various application domains, enabling practitioners and engineers to quickly grasp information in two- or three-dimensional (2D or 3D) maps \cite{ayesha2020overview}. A wide range of techniques for the visualization of high-dimensional data sets have been proposed. For example, Principal Component Analysis (PCA) is one of the prevalent techniques for reducing the dimensionality \cite{anaraki2014memory,pourkamali2017preconditioned}. The primary objective of PCA is to reduce dimensionality while preserving as much variance as possible in the given data set, where each principal component is a weighted combination of all the features or attributes in the original data set. One disadvantage of using PCA for visualization purposes is that data samples tend to be densely clustered together \cite{wu2017visualization}. On the other hand, Linear Discriminant Analysis (LDA) maximizes the separation between classes while performing dimensionality reduction \cite{xanthopoulos2013linear}. However, a significant drawback of LDA is that the output dimension is necessarily less than the number of categories, which makes it unappealing for binary classification problems.

The t-Distributed Stochastic Neighbor Embedding (t-SNE) algorithm \cite{van2008visualizing,arora2018analysis} is another popular method for visualizing high-dimensional data sets by giving each sample a location in a 2D or 3D map. In essence, t-SNE minimizes the Kullback-Leibler (KL) divergence between two probability distributions: a distribution that measures pairwise similarities in the input high-dimensional space, and another distribution that represents pairwise similarities of the corresponding low-dimensional embedded samples. Therefore, t-SNE is a nonlinear dimension reduction technique capable of preserving the local structure of the given data. As a result, t-SNE has found applications in many high-impact applications, such as visualizing biological \cite{kobak2019art} and geological \cite{balamurali2016t} data sets. 

Motivated by recent applications of t-SNE, we propose to employ this powerful technique for visualizing scientific data sets that originate from advanced computer simulations. Specifically, we consider a binary data set stemming from computational models of earthquake ground motions in structural engineering \cite{hariri2018simplified,pourkamali2021neural}. The objective is to train binary classifiers to predict the severity of the damage, i.e., safe vs. failed simulations. However, unfortunately, interpreting trained classifiers in the high-dimensional input space comprising a multitude of features is a formidable task for engineers. Therefore, we conduct a comprehensive set of numerical experiments to demonstrate the effectiveness of t-SNE for scientific data visualization. In our experiments, the embedded data samples in a 2D map serve as the input for classification algorithms. Another contribution of this work is utilizing six different classifiers to enhance our understanding about the impact of dimensionality reduction on the performance of various classifiers. 
To the best of our knowledge, this paper is the first application of t-SNE in the field of structural and earthquake engineering, providing future research direction to guide academic and industry researchers. The broader impact of this paper can be related to the quantitative risk assessment in structures and infra-structures subjected to multi-hazard scenarios, such as earthquake, hurricane, and flooding.

Moreover, we systematically investigate the impact of class-imbalanced data on the resulting 2D embedding provided by t-SNE. When using data-driven approaches, the class imbalance issue (i.e., considerably more instances of one class than another) poses a significant challenge as most machine learning algorithms overlook samples from minority classes \cite{krawczyk2016learning,buda2018systematic,hajibabaee2021kernel}. For example, in our case study, the number of safe simulations greatly exceeds the number of failed simulations. Although there are many techniques for tackling the class imbalance problem for classification problems, the remaining question is to examine the interface between t-SNE and imbalanced data. Hence, we propose to use a popular technique, known as the Synthetic Minority Oversampling Technique or SMOTE \cite{chawla2002smote,fernandez2018smote}, for improving the performance of t-SNE. 

The remainder of the paper is outlined as follows. In Section \ref{sec:method}, we review the machine learning techniques used in this work to handle high-dimensional class-imbalanced data. In Section \ref{sec:exp}, we describe our experimental setup and different approaches we take. In Section \ref{sec:data}, we outline the scientific simulation model, which is our case study. Finally, we present our numerical experiments and discussions in Section \ref{sec:result}. 

\section{Methods}\label{sec:method}


\subsection{t-SNE}\label{sec:t-SNE}

Given the difficulty of visualizing data with more than two dimensions, a fundamental task involves analyzing and interpreting high-dimensional data sets. For the sake of improved visualization, the t-SNE \cite{van2008visualizing} algorithm is employed to reduce the dimensionality. This algorithm is a powerful nonlinear dimensionality reduction technique for the visualization of data sets containing hundreds or even thousands of dimensions in 2D and 3D maps (we focus on 2D maps in this work).

To be formal, let us consider a data set that contains $n$ samples $x_1,\ldots,x_n$ in $\mathbb{R}^d$. The goal is to find a low-dimensional embedding or map points $y_1,\ldots,y_n$ in $\mathbb{R}^2$. The idea is that if two points are close in the input space, their corresponding map points should be close too. To this end, we define a conditional similarity between pairs of samples:
\begin{equation}
    p_{j|i} = \frac{\exp(-\|x_i-x_j\|_2^2/2\sigma_i^2)}{\sum_{k\neq i} \exp(-\|x_i-x_k\|_2^2/2\sigma_i^2)}. \nonumber
\end{equation}
The variance parameter $\sigma_i^2$ is different for each sample in the data set. We can also define a symmetric similarity measure as: $p_{ij}=(p_{i|j}+p_{j|i})/(2n)$. Next, we should define a new similarity measure for the mapping, denoted by $f$:
\begin{equation}
q_{ij}=\frac{f(\|x_i-x_j\|_2)}{\sum_{k\neq i} f(\|x_i-x_k\|_2)}.\nonumber
\end{equation}
The final ingredient of t-SNE is to ensure that these two metrics are close to each other based on the Kullback-Leiber (KL) divergence between the two distributions $p_{ij}$ and $q_{ij}$: 
\begin{equation}
    \text{KL}\big(P||Q\big)=\sum_{i,j} p_{ij}\log(p_{ij}/q_{ij}). \nonumber
\end{equation}

In the t-SNE algorithm, \textit{perplexity} is one of the most critical hyperparameters to be adjusted by the user. The perplexity parameter in t-SNE serves as a knob that sets the number of influential nearest neighbors, thus significantly impacting the resulting mapping of the input data. The standard range for the perplexity parameter is typically between 5-100, requiring a grid search for hyperparameter optimization.




\subsection{SMOTE}\label{sec:SMOTE}

The Synthetic Minority Oversampling Technique (SMOTE) \cite{chawla2002smote} is arguably the most popular data-level method for handling the class imbalance problem. 
The key idea is to generate new artificial samples to increase the size of the minority class. Hence, SMOTE makes learning from the minority class easier by better aligning the decision boundary toward an appropriate acknowledgment of all categories \cite{kovacs2019smote}.

\section{Experimental Settings}\label{sec:exp}

\begin{figure*}[ht!]
	\centering
	\includegraphics[width=0.95\textwidth]{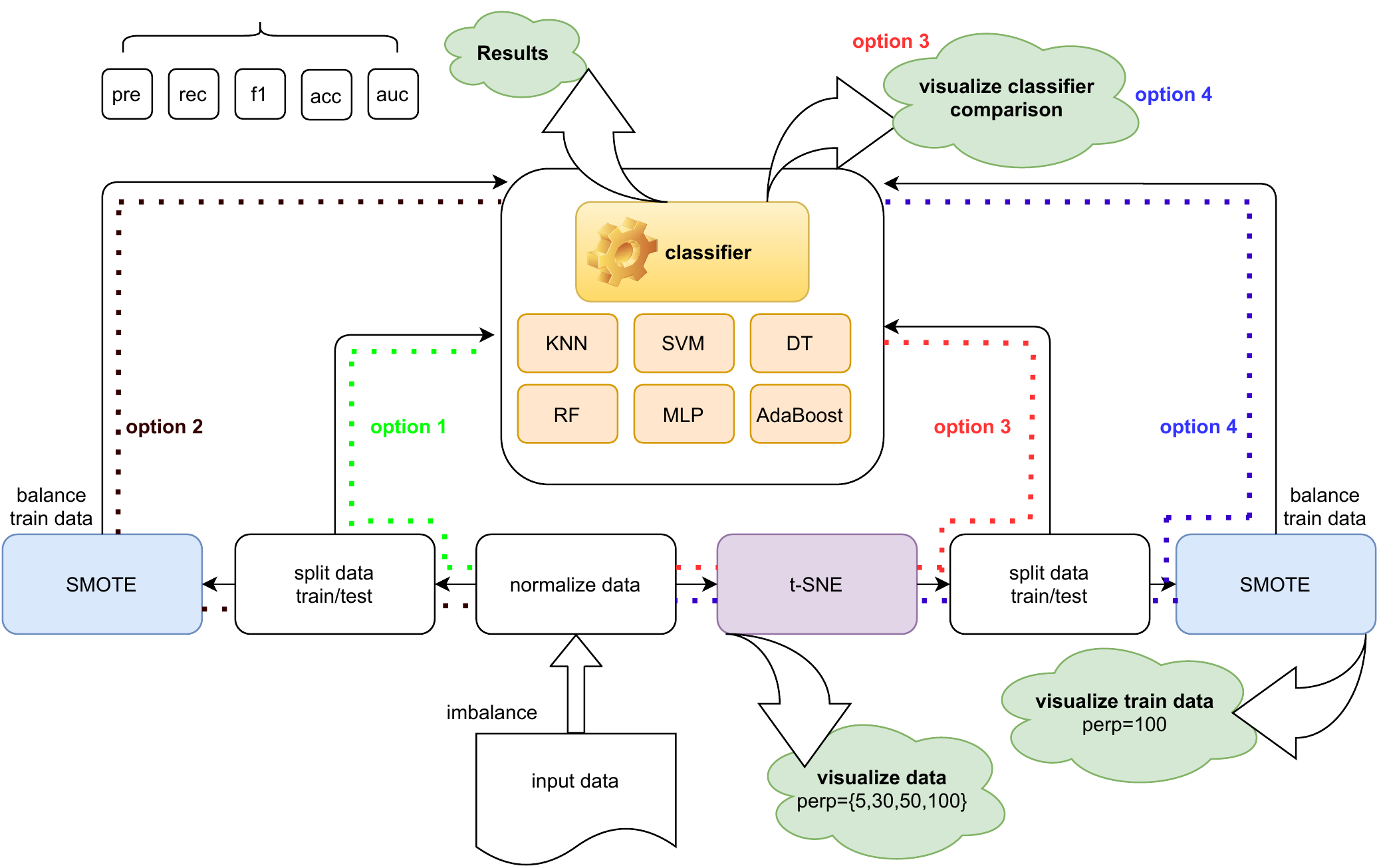}
	\caption{The modular experimental setting with the flow from the raw input scientific data set to results (visualization/classification).}
	\label{fig:experiment}
\end{figure*}

In this section, we outline four approaches that we take to systematically investigate the integration of t-SNE and SMOTE for analyzing scientific data. Fig.~\ref{fig:experiment} depicts a graphical overview of these four approaches, discussed in the following.

\subsection{Option 1}\label{sec:option1}

In this approach, we work with the original input space without further actions to tackle the class imbalance problem and to visualize the high-dimensional data. We normalize the data and then divide it into train/test sets. Finally, we train different classification algorithms, described in Section \ref{sec:classification}, while using the test set to evaluate their performance.

\subsection{Option 2}\label{sec:option2}

This is similar to option 1 with a minor difference. After splitting the data into train/test sets, we apply SMOTE to the training data set to generate synthetic data for balancing the number of samples. Note that we split our data set before using any over/under-sampling methods. Otherwise, we will bias our model, which is simply wrong because we are introducing data points for our future test set that do not exist. 

\subsection{Option 3}\label{sec:option3}

Here, we aim to visualize the data before using oversampling methods and classifiers to better understand the anatomy and structure of the data. To achieve this goal of visualization, dimensionality reduction is required. To this end, after normalizing the input data, we apply t-SNE which is commonly used for visualization purposes in 2D scatter plots. Then, we split the data and fit our classifiers to get the performance results. The interesting point of this approach is that, thanks to the dimensionality reduction step, we can easily have a visual comparison between classifiers in the final step.

\subsection{Option 4}\label{sec:option4}

Similar to option 3, we first apply t-SNE to the normalized data, and we then split it into train/test sets. At this point, we use SMOTE to balance the training data set. We visualize the balanced training set and the resulting classifiers.

\subsection{Classification Algorithms}\label{sec:classification}

In this study, we incorporate six classifiers for the binary classification task. Our classifiers are: i) K-Nearest Neighbors (KNN), ii) Radial Basis Function (RBF) Support Vector Machine (SVM), iii) Decision Tree (DT), iv) Random Forest (RF), v) Multi-Layer Perceptron (MLP), and vi) AdaBoost.
\begin{itemize} 

\item \textbf{K-Nearest Neighbors} is a type of instance-based learning: it does not attempt to construct a general internal model, but simply stores instances of the training data. Classification is computed based on a  majority vote of the nearest neighbors of each point: a query point is assigned the data class which has the most representatives within the nearest neighbors of the point. \texttt{scikit-learn} implements \texttt{KNeighborsClassifier}, where $k$ is an integer value specified by the user to define the number of nearest neighbors. In this work, we choose $k = 3$.

\item \textbf{RBF Support Vector Machine} finds the dividing hyperplane that separates both classes of the training set with the maximum possible margin. Then, the predicted label of a new, unseen data point is determined based on which side of the hyperplane it falls \cite{cite37}.
When training SVM with the radial basis function kernel, two parameters must be considered: $C$ and $\gamma$. In RBF kernel function, $\exp(-\gamma\|x_i-x_j\|^2)$, $\gamma$ must be greater than $0$.
The parameter $C$, common to all SVM kernels, trades off misclassification of training examples against simplicity of the decision surface. A low $C$ makes the decision surface smooth, while a high $C$ aims at classifying all training examples correctly as much as possible. $\gamma$ defines how much influence a single training example has. The larger $\gamma$ is, the closer other examples must be to be affected. We consider $(\gamma = 2, C = 1)$ in our model.
    
\begin{figure*}[ht!]
\centering
\includegraphics[width=0.96\textwidth]{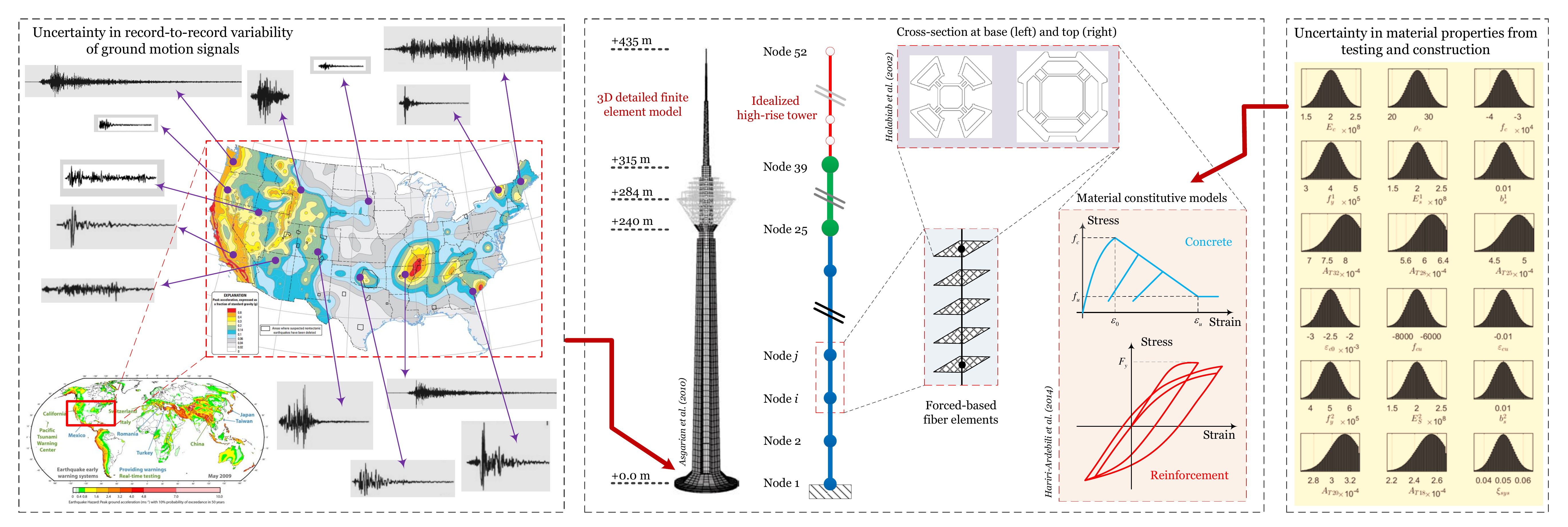}
\caption{Developed numerical model of a complex structural system with combined uncertainty inputs (schematic plot): uncertainty in applied stochastic ground motion records, as well as randomness in the simulated material/modeling properties.}
\label{fig:FE_model}
\end{figure*}


\item \textbf{Decision Trees} are a non-parametric learning method which employ a straightforward idea to solve the classification problem. Trees are built beginning with the tree's root and continuing down to its leaves. DT poses a set of carefully constructed questions with which a classification rule is developed through a set of attributes \cite{befcite53}. DT algorithms use a heuristic method or greedy strategy to direct their quest in the vast space of hypotheses for simplifying decision trees \cite{cite53}. We consider the maximum depth of the tree \texttt{max\_depth} = 5.

\item \textbf{Random Forest} builds multiple trees randomly among subsets of features to form branches of decision trees. Several trees are trained in random forests, instead of training a single tree \cite{cite56}. Bagging method is applied to build each decision tree, and then all the decision trees are combined to form the RF. We consider \texttt{max\_depth} = 5, \texttt{n\_estimators} = 10, and \texttt{max\_features} = 1.

\item \textbf{MLPClassifier} implements a multi-layer perceptron algorithm using backpropagation. MLP, a feedforward artificial neural network model, is a supervised learning algorithm. Our \texttt{MLPClassifier} model optimizes the log-loss function using \texttt{adam}, since this solver works pretty well on relatively large data sets (with thousands of training samples or more) in terms of both training time and validation score \cite{scikit-learn}. In our single hidden layer MLP model, we consider 
the regularization term parameter \texttt{alpha}=1 and the number of epochs \texttt{max\_iter}=1000.

\item \textbf{AdaBoost} is  an ensemble learning method based on creating powerful classifiers by combining weak learners. AdaBoost is an iterative algorithm in which, at any iteration, each weak classifier is trained, and the weight is assigned to, based on the accuracy achieved \cite{cite58}. Classifiers with higher accuracy levels are assigned with higher weights to have a more impact on the final prediction.

\end{itemize}

In this work, all algorithms under comparison are implemented in Python. To perform the classification task and training binary classifiers, we use built-in estimators in the \texttt{scikit-learn} package (Version 0.24.1). 
We use these functions with default arguments unless otherwise stated. 

\section{Case Study}\label{sec:data} 

Risk-based safety management of infra-structures is a critical task that involves various uncertainty sources. According to the performance-based earthquake engineering (PBEE) framework proposed by the Pacific Earthquake Engineering Research (PEER) center, any realistic risk analysis should consider aleatory and epistemic uncertainties. The former originates from uncertainties in the applied load (in this case, the ground motion record-to-record variability), while the latter has its roots in material and modeling randomness. Multiple studies reported the hybrid impact of these uncertainty sources \cite{helton1996guest, CelikEllingwood2010, jiang2017probability, hariri2019efficient, chen2019compatible}. While the concept of uncertainty quantification in risk-based assessment can be studied from different viewpoints, in connection to the proposed methodology in this paper, we aim to answer an important question: how to classify safe/failure events in the case of imbalanced data?

For this purpose, we consider a high-rise telecommunication tower as a case study \cite{HaririSamaniMirtaheri2014}. The height is over $400$ meters, made of reinforced concrete. The concrete shaft is the main load-carrying structure of the tower that transfers the lateral and gravitational loads to the foundation. We consider several modeling aspects, including material nonlinearities (i.e., cracking, crushing, and damage), and geometric nonlinearities. 
To reduce the computational burden, We develop a 2D model of the tower, including the head structure, shaft, and transition. A total of $10$ random models are generated using Latin Hypercube Sampling (LHS) to consider the variability of $18$ material/modeling parameters (i.e., concrete, steel, and system level damping). 
Probability density functions for all these $18$ variables are illustrated in Fig.~\ref{fig:FE_model} (right box).

Moreover, $100$ ground motions are selected world-wide to consider the aleatory uncertainty source. The record selection process is also random in order to cover a wide range of potential events. Fig.~\ref{fig:FE_model} (left box) illustrates a few selected records within the United States based on the national seismic hazard maps. Since as-recorded (i.e., unscaled) ground motion records are typically weak and may not cause failure to the structure, a scaling technique is used to generate other ground motions that are more destructive. Therefore, $700$ new records are generated (based on the original 100 ones), resulting in $800$ ground motion records that are used in our numerical model. While there are several optimal design of experiments to combine $10$ structural models from LHS with $800$ ground motion records \cite{cavazzuti2012optimization, hariri2018mcs}, we choose to use Randomized Complete Block Design, which is simply a full combination of all possible cases ($8,\!000$ events). Each event is a unique combination of a random model and a random ground motion.


\begin{figure*}[ht!]
	\centering
	\includegraphics[width=0.92\textwidth]{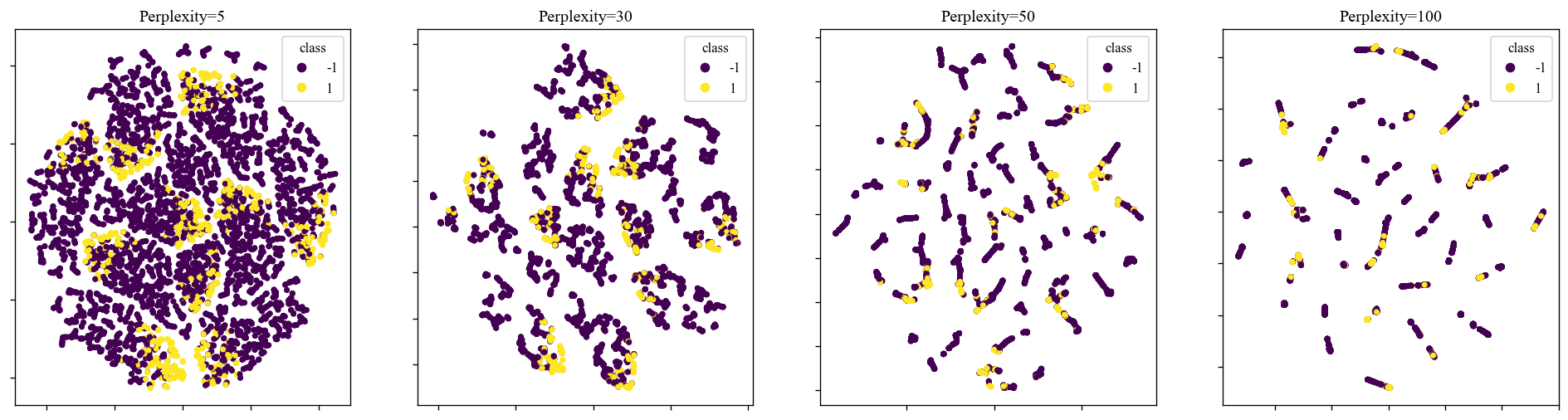}
	\caption{2D maps attained by t-SNE on the imbalanced scientific data set for prep={ 5, 30, 50, 100}.}
	\label{fig:t-SNEimb}
\end{figure*}

As discussed earlier, each random model has $18$ unique attributes from material/modeling (which are in the form of real numbers), and a time-dependent stochastic ground motion signal. Hence, we should convert the stochastic nature of the ground motion signal into a manageable number of scalar features. Such a conversion is already discussed in \cite{HaririSaoumaCollapseFragility, hariri2019series} for generic infrastructural systems. For each ground motion, we extract $31$ intensity measure (IM) parameters (or meta-features), including all peak values, as well as the intensity-, frequency-, and duration-dependent parameters. 

Combining the $31$ meat-features extracted from ground motion signals and the $18$  material/modeling attributes leads to $d=49$ features for each event or data sample. Therefore, the input data set consists of $n=8,\!000$ samples in $\mathbb{R}^{49}$. The outputs of our finite element simulation model provide the ground-truth labels, containing $-1$ for safe and $+1$ for failed simulations. To be precise, we have $n_-=6,\!937$ and $n_+=1,\!063$  (i.e., about $13\%$ of simulations are failed).


\section{Experimental Results and Discussion}\label{sec:result}

This section reports the results of our empirical investigation given the experimental setup described in Section~\ref{sec:exp}, which is also depicted in Fig~\ref{fig:experiment}. 

First, we plot 2D maps attained by the t-SNE algorithm for varying values of the perplexity parameter. As we observe in Fig.~\ref{fig:t-SNEimb}, when the perplexity parameter is set to $5$, the resulting 2D map is not useful because the two classes are very close to each other. However, as we increase this parameter, we notice a larger separation between the two classes of safe and failed simulations, thus t-SNE offers a practical visualization technique for the discussed scientific data.
Moreover, the reduced dimensionality data samples can be provided as inputs to the classification algorithms for developing prediction models.

In the second part of this section, we report several evaluation metrics for classification problems. These metrics are listed in the following: 
\begin{itemize}
\item \textbf{Precision} is defined as the ratio TP / (TP + FP), where TP is the number of true positives and FP denotes the number of false positives.
\item \textbf{Recall} is defined as the ratio TP / (TP + FN), where FN refers to false negatives. 
\item \textbf{F1-Score} combines the precision and recall scores.
\item \textbf{Balanced accuracy} avoids inflated performance estimates on class-imbalanced data. It is defined as the average of recall obtained on each class or the arithmetic mean of sensitivity (true positive rate) and specificity (true negative rate) in the following form:
\begin{equation}
   \text{balanced accuracy} = \frac{1}{2}\big(\frac{\text{TP}}{\text{TP} + \text{FN}} + \frac{\text{TN}}{\text{TN} + \text{FP}}\big).\nonumber
\end{equation}
Balanced accuracy takes values between $0$ and $1$, and higher values indicate better prediction models.
\item \textbf{AUC-Score} (Area Under Receiver-operating Characteristic Curve) score is a widely used metric for assessing performance in imbalanced learning \cite{koziarski2017ccr}. 
\end{itemize}

\begin{table*}[th!]
\caption{Reporting performance metrics using six classifiers and four options.} 
\centering 
\resizebox{\textwidth}{!}{\begin{tabular}{l|llllllllllllllllllll}
\cline{2-21}
 &
  \multicolumn{5}{c|}{\textit{\textbf{option 1}}} &
  \multicolumn{5}{c|}{\textit{\textbf{option 2}}} &
  \multicolumn{5}{c|}{\textit{\textbf{option 3}}} &
  \multicolumn{5}{c|}{\textit{\textbf{option 4}}} \\ \hline
\multicolumn{1}{|c|}{\textit{\textbf{classifier}}} &
  \multicolumn{1}{c}{\textbf{pre}} &
  \multicolumn{1}{c}{\textbf{rec}} &
  \multicolumn{1}{c}{\textbf{f1}} &
  \multicolumn{1}{c}{\textbf{acc}} &
  \multicolumn{1}{c}{\textbf{auc}} &
  \multicolumn{1}{c}{\textbf{pre}} &
  \multicolumn{1}{c}{\textbf{rec}} &
  \multicolumn{1}{c}{\textbf{f1}} &
  \multicolumn{1}{c}{\textbf{acc}} &
  \multicolumn{1}{c}{\textbf{auc}} &
  \multicolumn{1}{c}{\textbf{pre}} &
  \multicolumn{1}{c}{\textbf{rec}} &
  \multicolumn{1}{c}{\textbf{f1}} &
  \multicolumn{1}{c}{\textbf{acc}} &
  \multicolumn{1}{c}{\textbf{auc}} &
  \multicolumn{1}{c}{\textbf{pre}} &
  \multicolumn{1}{c}{\textbf{rec}} &
  \multicolumn{1}{c}{\textbf{f1}} &
  \multicolumn{1}{c}{\textbf{acc}} &
  \multicolumn{1}{c}{\textbf{auc}} \\ \hline
\multicolumn{1}{|l|}{\textbf{KNN}} &
  0.92 &
  0.93 &
  0.92 &
  0.82 &
  0.81 &
  0.92 &
  0.9 &
  0.91 &
  0.88 &
  0.88 &
  0.91 &
  0.91 &
  0.91 &
  0.8 &
  0.8 &
  0.91 &
  0.88 &
  0.89 &
  0.85 &
  0.9 \\ \hline
\multicolumn{1}{|l|}{\textbf{SVM}} &
  0.93 &
  0.93 &
  0.93 &
  0.83 &
  0.83 &
  0.93 &
  0.91 &
  0.92 &
  0.88 &
  0.88 &
  0.9 &
  0.91 &
  0.91 &
  0.78 &
  0.78 &
  0.92 &
  0.89 &
  0.9 &
  0.85 &
  0.9 \\ \hline
\multicolumn{1}{|l|}{\textbf{DT}} &
  0.92 &
  0.92 &
  0.92 &
  0.81 &
  0.81 &
  0.93 &
  0.91 &
  0.92 &
  0.9 &
  0.9 &
  0.85 &
  0.87 &
  0.86 &
  0.6 &
  0.6 &
  0.89 &
  0.65 &
  0.7 &
  0.65 &
  0.6 \\ \hline
\multicolumn{1}{|l|}{\textbf{RF}} &
  0.92 &
  0.93 &
  0.92 &
  0.82 &
  0.82 &
  0.93 &
  0.87 &
  0.89 &
  0.9 &
  0.9 &
  0.87 &
  0.88 &
  0.84 &
  0.6 &
  0.6 &
  0.89 &
  0.64 &
  0.7 &
  0.7 &
  0.7 \\ \hline
\multicolumn{1}{|l|}{\textbf{MLP}} &
  0.93 &
  0.93 &
  0.93 &
  0.81 &
  0.81 &
  0.93 &
  0.89 &
  0.9 &
  0.9 &
  0.9 &
  0.88 &
  0.88 &
  0.82 &
  0.5 &
  0.5 &
  0.91 &
  0.83 &
  0.86 &
  0.89 &
  0.9 \\ \hline
\multicolumn{1}{|l|}{\textbf{AdaBoost}} &
  0.94 &
  0.94 &
  0.94 &
  0.85 &
  0.85 &
  0.94 &
  0.92 &
  0.93 &
  0.92 &
  0.92 &
  0.86 &
  0.87 &
  0.84 &
  0.53 &
  0.53 &
  0.9 &
  0.8 &
  0.83 &
  0.81 &
  0.8 \\ \hline
\end{tabular}}
\label{table:result} 
\end{table*}

Table~\ref{table:result} reports these performance metrics when applying the six introduced classifiers to our scientific data set. According to this table, using SMOTE, improves balanced accuracy (acc) and AUC scores across all four options. Although this behavior is expected in the high-dimensional input space, we have shown similar trends when using t-SNE for visualization purposes. As can be seen from Table~\ref{table:result}, although option 2 achieved the highest balanced accuracy (acc) and AUC scores, we lose the opportunity for the visualization of the data's structure and classifier's decision surface to compare the performance of various classification algorithms beyond standard numerical metrics. 

Furthermore, we notice that using SMOTE when working with 2D maps is crucial. For example, comparing option 3 with 4, the balanced accuracy score (acc) increases from $0.5$ to $0.89$ for MLP. Additionally, comparing option 2 with 4, balanced accuracy (acc) and AUC scores seem almost the same across all classifiers except DT and RF. Therefore, it is wise to choose option 4 since we can take advantage of visualization with no significant loss in performance. 
Looking at the Table~\ref{table:result}, the same rule applies to all performance metrics reported in this study.  In this experimental investigation, we discussed balanced accuracy (acc) and AUC scores because they are considered the most critical performance metrics in imbalance learning.
Also, it is interesting to notice that the K-Nearest Neighbors (KNN) classifier performs well in option 3 and 4 despite its simplicity. 
However, if we have to pick a single classifier in this experiment, MLP achieves the highest value of balanced accuracy among the six selected classifiers in option 4.

\begin{figure*}[ht!]
	\centering
	\includegraphics[width=0.8\textwidth]{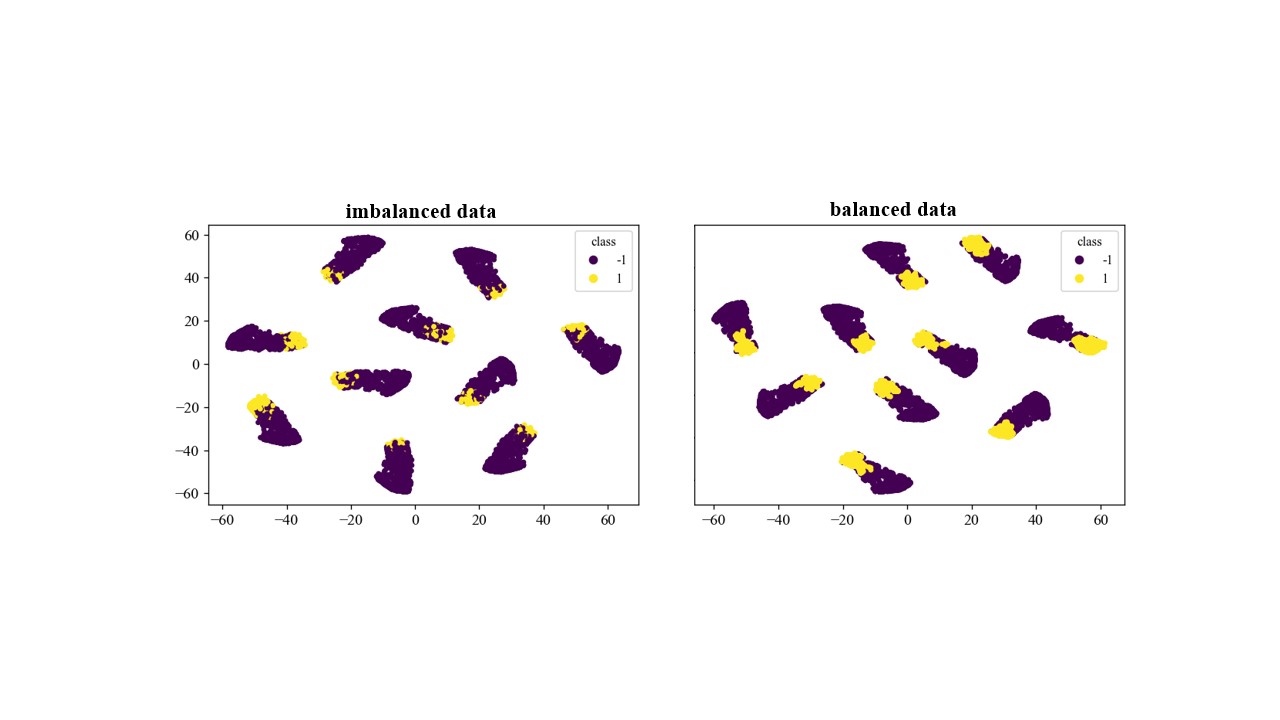}
	\vspace{-22mm}
	\caption{2D maps attained by applying t-SNE to the training data set (perp = 100) using option 3 vs.~option 4.}
	\label{fig:perp100}
\end{figure*}

As depicted in Fig.~\ref{fig:experiment}, we can take advantage of data visualization in option 3 and 4, where we utilize the t-SNE algorithm. To see how SMOTE works to handle the imbalance nature of the scientific data in our case study, Fig.~\ref{fig:perp100} shows 2D plots of the training data set. In this figure, purple color shows the safe simulations whereas the yellow color represents failed ones, i.e., the minority class in our imbalanced data set. As the right figure illustrates in Fig.~\ref{fig:perp100}, with the help of SMOTE, the re-sampled data set is well balanced. That is, 
the dense yellow region clearly shows that how oversampling methods (SMOTE here) is capable of generating synthetic data points to handle the class imbalance problem. As discussed earlier, to avoid bias in the model, SMOTE is only utilized on the training data set (instead of applying to the whole data set). Fig.~\ref{fig:perp100} presents the imbalanced/balanced training data sets using option 3 vs.~option 4, where the perplexity parameter is set to $100$. 

\begin{figure*}[ht!]
	\centering
	\includegraphics[width=0.89\textwidth]{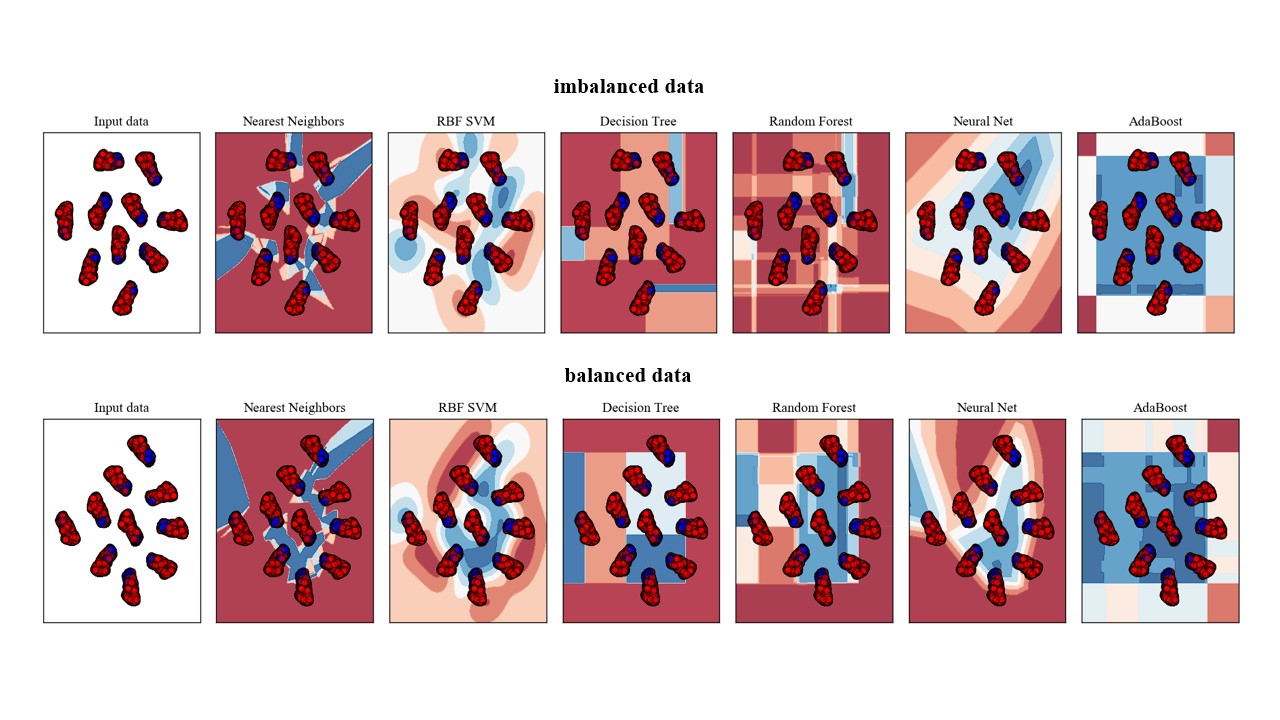}
	\vspace{-15mm}
	\caption{Classifier comparison using option 3 vs.~option 4.}
	\label{fig:comparison}
\end{figure*}

Finally, in Fig~\ref{fig:comparison}, we visualize and compare the trained classifiers using 2D plots attained by t-SNE with the perplexity value of $100$. In this figure, safe/failed simulations are represented by red and blue colors, respectively. Also training points are showed in solid colors while testing points in semi-transparent.
The point of this figure is to illustrate the nature of decision surface for each classifier. We apply six different classification algorithms to our given data set to visualize the decision functions. The bottom plot refers to option 4,  where we balance our training data using SMOTE, whereas the top plot is related to the imbalanced training data set, i.e., option 3. 
We discern that the neural network model or MLP, RBF SVM,  and K-Nearest Neighbors (KNN) on the balanced data set using SMOTE provide a meaningful decision surface to separate the two classes.  On the other hand, DT, RF, and AdaBoost fail to provide reasonable decision boundaries based on visualizing classifiers. This observation is consistent with the performance metrics reported in Table~\ref{table:result} because DT, RF, and AdaBoost ranked as three classifiers with the lowest balanced accuracy scores in option 4. \\
To conclude, we note that t-SNE offers a powerful technique for visualizing high-dimensional scientific data sets, allowing for inspecting prediction models beyond standard numerical performance metrics. Hence, our empirical investigation provides a roadmap for practitioners and engineers who want to better understand the performance of machine learning models in high-dimensional data regimes. Furthermore, we showed that integrating re-sampling techniques and t-SNE is essential for class-imbalanced data sets.

This work just focused on the binary classification problem. In our future work, we plan to visualize high-dimensional multi-class scientific data sets using different dimensionality reduction techniques such as t-SNE, PCA, and LDA and make a fair comparison. Furthermore, we can investigate how imbalanced learning strategies handle cases with more than three classes (e.g., ten or more categories).


\vspace{-3mm}

\bibliographystyle{IEEEtran}
\bibliography{ref.bib}

\end{document}